\documentclass[conference]{IEEEtran}
\IEEEoverridecommandlockouts

\usepackage{cite}
\usepackage{amsmath,amssymb,amsfonts}
\usepackage{algorithmic}
\usepackage{graphicx}
\usepackage{textcomp}
\usepackage{epsfig}
\usepackage{xcolor}
\usepackage{epstopdf}
\usepackage{caption}
\usepackage{subcaption}
\usepackage{booktabs} 
\usepackage[section]{placeins}
\usepackage{multirow}
\usepackage{url}
\usepackage[comma, numbers]{natbib}
\def\BibTeX{{\rm B\kern-.05em{\sc i\kern-.025em b}\kern-.08em
    T\kern-.1667em\lower.7ex\hbox{E}\kern-.125emX}}
\begin{document}

\title{A Decomposition Modeling Framework for Seasonal Time-Series Forecasting\\
}

\author{\IEEEauthorblockN{Yining Pang}
\IEEEauthorblockA{\textit{Hebei Finance University} \\
Baoding Hebei \\
p18132109069@163.com}
\and
\IEEEauthorblockN{Chenghan Li}
\IEEEauthorblockA{\textit{ZJU-UIUC Institute} \\
Haining, China\\
chenghan.20@intl.zju.edu.cn}}

\maketitle

\begin{abstract}
Seasonal time series exhibit intricate long-term dependencies, posing a significant challenge for accurate future prediction. This paper introduces the Multi-scale Seasonal Decomposition Model (MSSD) for seasonal time-series forecasting. Initially, leveraging the inherent periodicity of seasonal time series, we decompose the univariate time series into three primary components: Ascending, Peak, and Descending. This decomposition approach enhances the capture of periodic features. By addressing the limitations of existing time-series modeling methods, particularly in modeling the Peak component, this research proposes a multi-scale network structure designed to effectively capture various potential peak fluctuation patterns in the Peak component. This study integrates Conv2d and Temporal Convolutional Networks to concurrently capture global and local features. Furthermore, we incorporate multi-scale reshaping to augment the modeling capacity for peak fluctuation patterns. The proposed methodology undergoes validation using three publicly accessible seasonal datasets. Notably, in both short-term and long-term fore-casting tasks, our approach exhibits a 10$\%$ reduction in error compared to the baseline models.
\end{abstract}


\section{Introduction}
Seasonal time series, marked by recurring and cyclical patterns, play a crucial role in our daily lives. Traditional statistical methods, such as ARIMA and Holt-Winters \cite{li2021clothing,moorthy1988short} have utilized linear features for prediction. However, their accuracy is rather limited \cite{contreras2003arima,elmunim2013short} prompting the need for exploration of more sophisticated models. In contrast, Recurrent Neural Networks (RNNs) and their variations \cite{hochreiter1997long,chung2014empirical} have demonstrated enhanced performance by incorporating nonlinear features\cite{qin2017dual,lai2018modeling}. Despite their progress, these models encounter challenges related to computationally inefficient feature extraction, as early features gradually diminish while hidden states epvolve.

Mopdels based on transformers., spuch as Informer \citep{zhou2021informer} and Autoformer\citep{wu2021autoformer}, have emerged as leading contenders.  In the realm of time series, these models employ self-attention mechanisms to capture intricate sequence features, facilitating the comprehension of complex data relationships. This represents a significant paradigm shift, overcoming certain limitations inherent in traditional approaches and demonstrating remarkable predictive capabilities. Despite the progress achieved by transformer-based models, the challenge of computational overhead remains a noteworthy concern. Several optimization strategies have been implemented to mitigate this challenge; however, additional attention and enhancements are warranted. The pursuit of more efficient and scalable transformer architectures continues, with the goal of achieving a balance between computational efficiency and predictive accuracy.

Current efforts involve continual research on variants of transformer models, investigating innovative attention mechanisms, and developing strategies to optimize computational processes. The goal is not only to maintain or enhance predictive accuracy but also to ensure their scalability and practicality for real-world applications. To conclude, although traditional statistical methods and RNNs have established the foundation for time-series prediction, transformer-based models signify a substantial advancement. The self-attention mechanisms enable these models to capture intricate data relationships, yet the challenge of computational overhead persists. Continuous research and development in this field are imperative to refine transformer models, rendering them more efficient and practical across diverse applications, including the precise prediction of seasonal time-series datasets.

The convolution-based structures for time-series prediction, exemplified by Timesnet \citep{wu2022timesnet} and MICN \citep{wang2022micn}, have successfully reduced the time-memory overhead in series prediction. Nevertheless, these network structures have not fully considered the intricate periodic patterns inherent in seasonal time series. Currently, convolution-based models have achieved notable advancements in enhancing computational efficiency, effectively alleviating the time-memory burden associated with traditional methods and recurrent neural networks. However, when applied to seasonal time series, these models fall short of fully capturing their distinctive cyclical characteristics.

A significant portion of current research is focused on examining correlations between time steps through attention mechanisms, with the aim of extracting seasonal features but often neglecting the inherent richness of the periodic nature of seasonal sequences. To tackle these challenges, a multi-scale seasonal decomposition time-series modeling framework named MSSD is proposed. In this framework, we decompose univariate seasonal time series into three components, each representing distinct fluctuation trends: Ascending, peaking, and descending components. The Ascending and Descending components display fixed fluctuation patterns modeled using linear regression. To address the complex fluctuation patterns of the Peak component, a multi-scale convolutional network called SDNet is introduced, utilizing Conv2d and causal convolution to extract both local and global correlations. The final predictions are derived from the combination of diverse prediction outcomes. The key contributions of this paper can be summarized below:

\begin{itemize}[] 
    \item A decomposition forecasting framework for seasonal time series is proposed. By decomposing univariate seasonal time series according to the wave pattern, the interpretability of the model is enhanced
    \item In this framework, we propose a novel method based on multi-scale convolutional networks (SDNet) to extract local and global correlations using Conv2d and causal convolution.
\end{itemize}

The MSSD performance is verified on multiple real-world data sets that exceed SOTA models.

\section{Methodology}
\begin{figure*}[htbp] 
  \centering
  \includegraphics[width=1\textwidth]{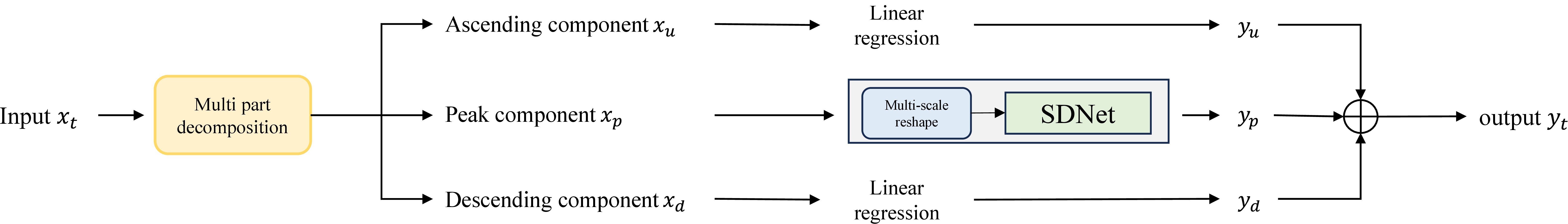} 
  \caption{Overall architecture of MSSD.}
  \label{fig:model}
\end{figure*}

Fig. \ref{fig:model} illustrates the overall structure of MSSD. Drawing inspiration from the periodicity inherent in time-series data, this paper introduces a module for decomposing time-series trend patterns. Subsequently, we employ distinct modules to predict the Ascending, Peak, and Descending components individually. Then, we aggregate the individual prediction results to obtain the final prediction, denoted as $y_{t}$. Further details will be provided in the subsequent sections (refer to Fig. \ref{fig:model}).

\subsection{Decomposition module}
Previous studies frequently employed one-dimensional convolution-based methods for time-series decomposition\cite{bai2018empirical}.  Previous research has consistently identified the standard period of seasonal cycle time series as typically 24 hours\cite{box1978analysis}. In the methodology employed in this paper, we categorize the daily cycle sequence into ascending, peak, and descending phases based on the average score method. For a comprehensive understanding, refer to formulas (1) through

\begin{equation}
T = 24 \cdot i \label{eq:time}
\end{equation}
\begin{equation}
x_u = x[0:\frac{T}{3}],\quad x_p = x[\frac{T}{3}: \frac{2T}{3}],\quad x_d = x[\frac{2T}{3}:T] \label{eq:parts}
\end{equation}
\begin{equation}
x = x_u + x_p + x_d \label{eq:sum}
\end{equation}

\begin{figure}[htbp] 
  \centering
  \includegraphics[width=0.5\textwidth]{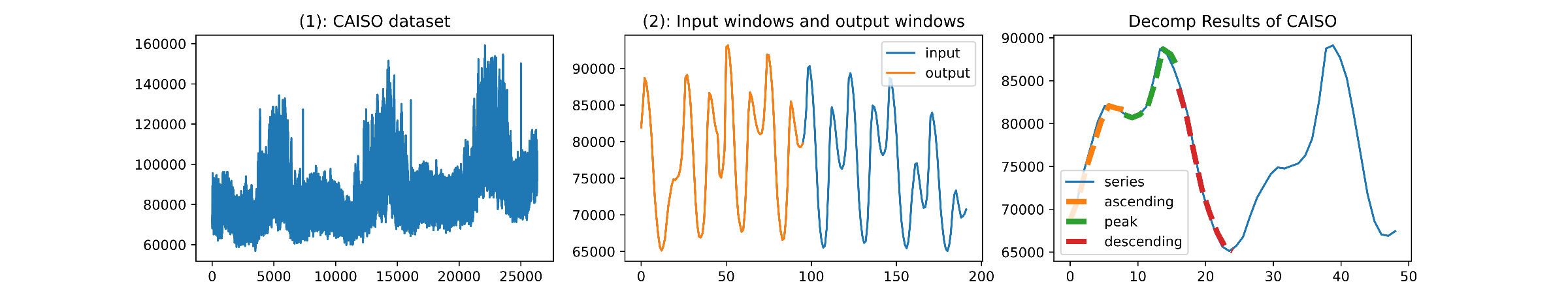} 
  \captionsetup{justification=centering}
  \caption{the specifics of the CAISO dataset. (1) represents the CAISO dataset itself; (2) delineates the windows for input and output series; and (3) showcases the decomposition results of CAISO.}
  \label{fig:caiso}
\end{figure}

In the given formula, $i$ denotes the sampling interval of the dataset, where, for example, 1 hour is represented as 1 and 15 minutes as 4. The variable $x$ represents a time series within a 24-hour period. The details of visualizations for the CAISO datasets are illustrated in Fig.\ref{fig:caiso}.

\subsection{Linear Regression module}\label{subsec2}
In this study, we utilize simple linear regression to model both ascending and descending components. As illustrated in Fig.\ref{fig:caiso}, seasonal time series display relatively stable fluctuation trends in ascending and descending components. Using simple regression for modeling purposes can reduce computational costs and improve the interpretability of the model.
\begin{equation}
y_u = \text{linear regression}(x_u) \label{eq:regression_u}
\end{equation}
\begin{equation}
y_d = \text{linear regression}(x_d) \label{eq:regression_d}
\end{equation}

\begin{figure}[!htbp] 
  \centering
  \includegraphics[width=0.5\textwidth]{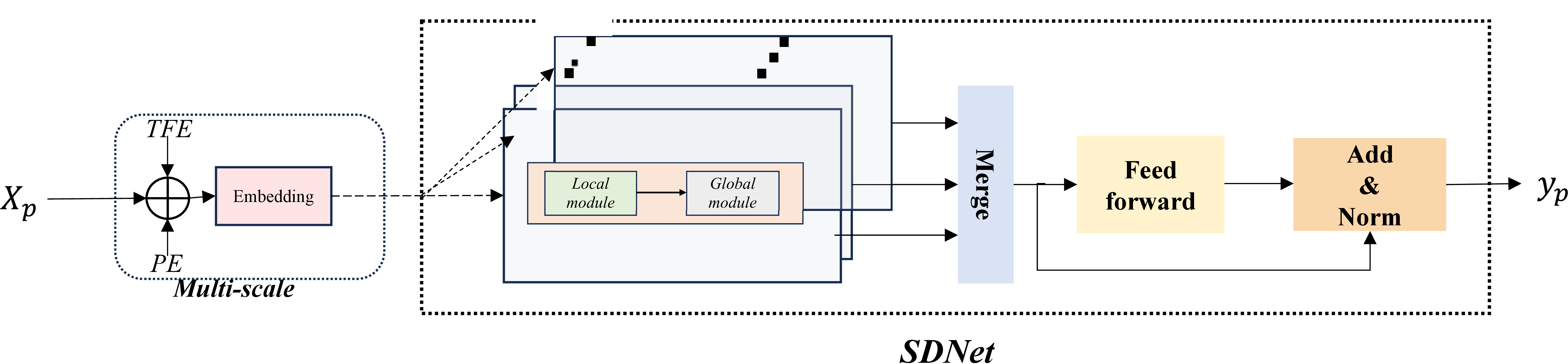} 
  \captionsetup{justification=centering}
  \caption{Seasonal component prediction module.}
  \label{fig:SDNet}
\end{figure}

As depicted in Fig.~\ref{fig:SDNet}, the SDNet prediction network primarily models the intricate peak fluctuation segment. Multiple multi-scale operations are applied to the input sequence  $x_{p}$ to capture local features and global correlations. Following this, the results from the distinct branches are aggregated to integrate the information within the sequence. This process can be summarized as follows:
\begin{equation}
\begin{aligned}
\{ x_{p,1}, \ldots, x_{p,l} \} &= \text{Multi-head}(x_p) \\
y_{p,i} &= \text{SDNet}(x_{p,i}) \\
y_p &= \text{Concat}(y_{p,i})
\end{aligned}
\label{eq:transformed_eq}
\end{equation}

\textbf{Multi-scale}: Seasonal periodic series usually contain complex features. In order to fully tap into these features, we are inspired by the work of multi-head attention modules. From Eq.~\eqref{eq:transformed_eq}, $l$ means the number of heads.

\begin{figure}[!htbp] 
  \centering
  \includegraphics[width=0.5\textwidth]{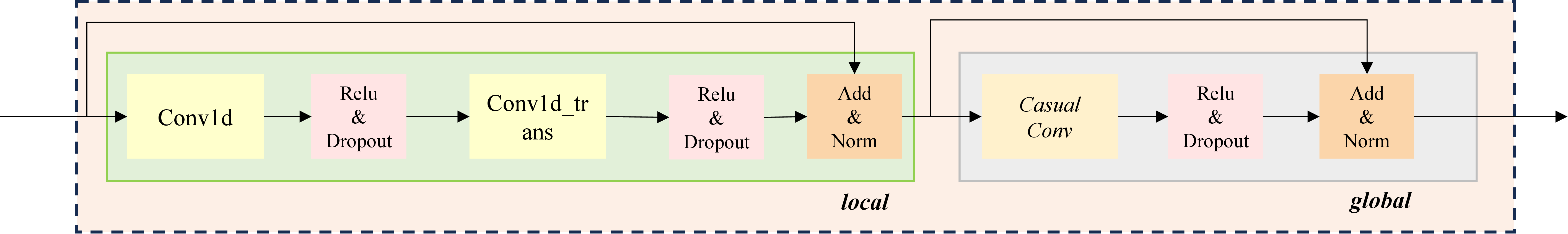} 
  \captionsetup{justification=centering}
  \caption{Local-Global convolution block architecture.}
  \label{fig:local-global}
\end{figure}

\textbf{Local-Global convolution block architecture}: SDNet consists of multiple branches, each handling distinct multi-scale results to simulate potentially diverse temporal patterns. As depicted in Fig.~\ref{fig:local-global}, the local-global module extracts both local features and global dependencies in each branch. Specifically, the local modules use one-dimensional convolution to facilitate the learning of similar wave features. This process includes the following steps:
\begin{equation}
x_{p,i}^{local, i} = \text{Norm}(\text{Conv1d}(x_{p,i})), \quad \text{for } i_{\text{kernel}=i}
\label{eq:local_conv}
\end{equation}

In the case of Conv1d, we configure stride $=$ kernel $= i$, effectively compressing local features. $x_{p,i}^{local, i}$ signifies the outcome of local feature compression, resulting in a brief sequence.

\textbf{Dilated Causal Convolutions}: Dilated convolution increases the receptive field by injecting holes into standard convolution. In this case, we use parameter $d$ to represent the sampling rate. A low-level $d=1$ implies sampling each point in the input, while an intermediate-level $d=2$ means sampling every 2 points in the input. Intuitively, dilated convolution results in an exponentially growing effective window size with the number of layers.


\section{Experiment}

\subsection{Experimental Setup}

\textbf{Dataset}: In this research, we present a thorough overview of the experimental datasets. Table \ref{tab:dataset_details} shows the details of three public datasets.
\begin{table}[!htbp]
  \centering\resizebox{\linewidth}{!}{
  \begin{tabular}{lllllll}
    \toprule
    Dataset & Source & Len & dim & Freq \\
    \midrule
    Electricity & \url{https://archive.ics.uci.edu/dataset/321/electricityloaddiagrams20112014} & 26304 & 322 & 1h \\
    Traffic & \url{http://pems.dot.ca.gov/} & 17544 & 863 & 1h \\
    CAISO & \url{https://www.caiso.com/Pages/default.aspx} & 13421 & 1 & 1h \\
    \bottomrule
  \end{tabular}}
  \caption{Details of Public Datasets}
  \label{tab:dataset_details}
\end{table}

\textbf{Baselines}: This investigation employs four advanced transformer models, specifically FEDformer~\cite{zhou2022fedformer}, Autoformer~\cite{wu2021autoformer}, Informer~\cite{zhou2021informer}, and LogTrans~\cite{li2019enhancing}, as integral components of our baseline models. Furthermore, we incorporate two enhanced RNN models---LSTM\cite{hochreiter1997long} and LSTNet~\cite{lai2018modeling}---alongside the CNN-based model MICN.

\textbf{Univariate results}: Table~\ref{tab:univariate_results_1} presents the outcomes of univariate time-series forecasting, highlighting the remarkable success of MSSD. Significantly, in short-term forecasting tasks using the CAISO dataset, MSSD exhibits a noteworthy average reduction of 3.8\% in MAE (6.3\%/2.7\%/2.6\%) compared to MICN. Likewise, MSSD demonstrates a substantial average reduction of 36.6\% (26.7\%/36.7\%/48.9\%) and -1.67\% (-4.7\%/8.3\%/-9.1\%) in MAE for the electricity dataset. In long-term forecasting tasks, it is evident that MSSD outperforms most error metrics. For multivariate time series analysis, MSSD was independently tested for each variable.

\begin{table}[h]
  \centering\resizebox{\linewidth}{!}{
  \begin{tabular}{lccccccc}
    \toprule
    & & \multicolumn{2}{c}{CAISO} & \multicolumn{2}{c}{Electricity} & \multicolumn{2}{c}{Traffic} \\
    \cmidrule(lr){3-4} \cmidrule(lr){5-6} \cmidrule(lr){7-8}
    Models & metrics & MSE & MAE & MSE & MAE & MSE & MAE \\
    \midrule
    LogTrans & 24 & 0.184 & 0.334 & 0.285 & 0.352 & 0.196 & 0.319 \\
             & 48 & 0.205 & 0.327 & 0.301 & 0.374 & 0.209 & 0.321 \\
             & 96 & 0.199 & 0.343 & 0.288 & 0.393 & 0.226 & 0.317 \\
    \cmidrule{1-8}
    Informer & 24 & 0.192 & 0.31 & 0.455 & 0.519 & 0.217 & 0.349 \\
             & 48 & 0.196 & 0.337 & 0.457 & 0.526 & 0.26 & 0.305 \\
             & 96 & 0.212 & 0.32 & 0.484 & 0.538 & 0.257 & 0.353 \\
    \cmidrule{1-8}
    Autoformer & 24 & 0.191 & 0.327 & 0.325 & 0.409 & 0.204 & 0.311 \\
              & 48 & 0.194 & 0.35 & 0.327 & 0.406 & 0.229 & 0.306 \\
              & 96 & 0.204 & 0.353 & 0.341 & 0.438 & 0.246 & 0.346 \\
    \cmidrule{1-8}
    Fedformer & 24 & 0.142 & 0.309 & 0.227 & 0.374 & 0.184 & 0.262 \\
              & 48 & 0.147 & 0.311 & 0.241 & 0.361 & 0.196 & 0.287 \\
              & 96 & 0.154 & 0.306 & 0.253 & 0.37 & 0.207 & 0.312 \\
    \cmidrule{1-8}
    MICN & 24 & 0.105 & 0.237 & 0.267 & 0.341 & 0.192 & \textbf{0.314} \\
         & 48 & 0.112 & 0.257 & 0.304 & 0.354 & 0.167 & 0.329 \\
         & 96 & 0.106 & 0.269 & 0.31 & 0.398 & 0.158 & \textbf{0.241} \\
    \cmidrule{1-8}
    MSSD & 24 & \textbf{0.067} & \textbf{0.222} & \textbf{0.087} & \textbf{0.25} & \textbf{0.161} & 0.329 \\
         & 48 & \textbf{0.091} & \textbf{0.25} & \textbf{0.076} & \textbf{0.224} & \textbf{0.146} & 0.325 \\
         & 96 & \textbf{0.101} & \textbf{0.262} & \textbf{0.061} & \textbf{0.203} & \textbf{0.108} & 0.263 \\
    \bottomrule
  \end{tabular}}
  \caption{Univariate short and long-term series forecasting results, where input length \( I = 96 \) and prediction length \( O \in \{24, 48, 96\} \).}
  \label{tab:univariate_results_1}
\end{table} 

\begin{table}[htbp]
  \centering\resizebox{\linewidth}{!}{
  \begin{tabular}{lccccccc}
    \toprule
    & & \multicolumn{2}{c}{CAISO} & \multicolumn{2}{c}{Electricity} & \multicolumn{2}{c}{Traffic} \\
    \cmidrule(lr){3-4} \cmidrule(lr){5-6} \cmidrule(lr){7-8}
    Models & metrics & MSE & MAE & MSE & MAE & MSE & MAE \\
    \midrule
      LogTrans & 192 & 0.294 & 0.417 & 0.432 & 0.483 & 0.314 & 0.408 \\
               & 336 & 0.307 & 0.456 & 0.43 & 0.483 & 0.387 & 0.453 \\
               & 720 & 0.312 & 0.489 & 0.491 & 0.531 & 0.491 & 0.437 \\
      \midrule
      Informer & 192 & 0.461 & 0.629 & 0.557 & 0.558 & 0.299 & 0.376 \\
               & 336 & 0.504 & 0.617 & 0.636 & 0.613 & 0.312 & 0.387 \\
               & 720 & 0.499 & 0.606 & 0.819 & 0.682 & 0.366 & 0.436 \\
      \midrule
      Autoformer & 192 & 0.345 & 0.453 & 0.345 & 0.428 & 0.3 & 0.369 \\
                & 336 & 0.329 & 0.476 & 0.406 & 0.47 & 0.323 & 0.263 \\
                & 720 & 0.343 & 0.481 & 0.565 & 0.581 & 1.26 & 0.867 \\
      \midrule
      FEDformer & 192 & 0.216 & 0.329 & 0.282 & 0.386 & 0.336 & \textbf{0.205} \\
                & 336 & 0.209 & 0.321 & 0.346 & 0.431 & 0.219 & 0.323 \\
                & 720 & 0.196 & 0.307 & 0.422 & 0.484 & 0.339 & \textbf{0.244} \\
      \midrule
      MICN & 192 & 0.119 & 0.304 & 0.3 & 0.394 & 0.154 & 0.236 \\
           & 336 & 0.121 & 0.296 & 0.323 & 0.413 & 0.165 & \textbf{0.243} \\
           & 720 & 0.124 & 0.285 & 0.364 & 0.449 & 0.182 & 0.264 \\
      \midrule
      MSSD & 192 & \textbf{0.107} & \textbf{0.271} & \textbf{0.069} & \textbf{0.215} & \textbf{0.138} & 0.317 \\
           & 336 & \textbf{0.097} & \textbf{0.258} & \textbf{0.068} & \textbf{0.214} & \textbf{0.146} & 0.327 \\
           & 720 & \textbf{0.093} & \textbf{0.255} & \textbf{0.064} & \textbf{0.207} & \textbf{0.17} & 0.346 \\
      \bottomrule
    
  \end{tabular}}
  \caption{Univariate short and long-term series forecasting results, where input length \( I = 96 \) and prediction length \( O \in \{192, 336, 720\} \).}
  \label{tab:univariate_results_2}
\end{table} 

\textbf{Multivariate results}:In the domain of multivariate long-term series forecasting, MSSD stands out as a pinnacle of state-of-the-art performance across diverse benchmarks and prediction length configurations, supported by the comprehensive results in Table ~\ref{tab:multivariate_results}. Significantly, MSSD consistently outperforms its counterparts, including the formidable MICN, demonstrating its prowess with an impressive 18\% average reduction in MSE across two distinct datasets. This significant improvement in MSE substantiates MSSD's capacity to excel in addressing a spectrum of challenges in time-series forecasting within real-world seasonal time-series applications. These findings highlight MSSD as a robust and versatile solution with the potential to significantly enhance predictive accuracy across various forecasting scenarios.

\begin{table}[htbp]
  \centering
  \resizebox{\linewidth}{!}{
  \begin{tabular}{lcccccccccccccccc}
    \toprule
    Dataset & Metrics & \multicolumn{2}{c}{MSSD} & \multicolumn{2}{c}{MICN} & \multicolumn{2}{c}{FEDformer} & \multicolumn{2}{c}{Autoformer} & \multicolumn{2}{c}{Informer} & \multicolumn{2}{c}{LSTNet} & \multicolumn{2}{c}{LSTM} \\
    \cmidrule(lr){3-4} \cmidrule(lr){5-6} \cmidrule(lr){7-8} \cmidrule(lr){9-10} \cmidrule(lr){11-12} \cmidrule(lr){13-14} \cmidrule(lr){15-16}
    & & MSE & MAE & MSE & MAE & MSE & MAE & MSE & MAE & MSE & MAE & MSE & MAE & MSE & MAE \\
    \midrule
    \multirow{4}{*}{Electricity} & 96 & \textbf{0.149} & \textbf{0.284} & 0.164 & 0.269 & 0.193 & 0.308 & 0.201 & 0.3217 & 0.274 & 0.368 & 0.68 & 0.645 & 0.375 & 0.437 \\
    & 192 & \textbf{0.166} & \textbf{0.271} & 0.177 & 0.285 & 0.201 & 0.315 & 0.222 & 0.334 & 0.296 & 0.386 & 0.725 & 0.676 & 0.442 & 0.473 \\
    & 336 & \textbf{0.183} & \textbf{0.296} & 0.193 & 0.304 & 0.214 & 0.329 & 0.231 & 0.338 & 0.3 & 0.394 & 0.828 & 0.727 & 0.439 & 0.473 \\
    & 720 & \textbf{0.199} & \textbf{0.299} & 0.212 & 0.321 & 0.246 & 0.355 & 0.254 & 0.361 & 0.373 & 0.439 & 0.957 & 0.811 & 0.98 & 0.814 \\
    \midrule
    \multirow{4}{*}{Traffic} & 96 & \textbf{0.471} & \textbf{0.294} & 0.519 & 0.309 & 0.587 & 0.366 & 0.613 & 0.388 & 0.719 & 0.391 & 1.107 & 0.685 & 0.843 & 0.453 \\
    & 192 & \textbf{0.486} & \textbf{0.299} & 0.537 & 0.315 & 0.604 & 0.373 & 0.616 & 0.382 & 0.696 & 0.379 & 1.157 & 0.706 & 0.847 & 0.453 \\
    & 336 & \textbf{0.511} & \textbf{0.304} & 0.534 & 0.313 & 0.621 & 0.383 & 0.622 & 0.277 & 0.777 & 0.42 & 1.216 & 0.73 & 0.853 & 0.455 \\
    & 720 & \textbf{0.563} & \textbf{0.305} & 0.577 & 0.325 & 0.626 & 0.382 & 0.66 & 0.408 & 0.864 & 0.472 & 1.481 & 0.805 & 1.5 & 0.805 \\
    \bottomrule
  \end{tabular}}
  \caption{Multivariate series forecasting results, where the input length $I = 96$ and prediction length $O \in \{96, 192, 336, 720\}$}
  \label{tab:multivariate_results}
\end{table} 

\subsection{Model Analysis}
\textbf{Local-Global Module vs. Auto-correlation, Self-Attention:}
In this investigation, we introduced a Seasonal Decomposition Network (SDNet) featuring a convolutional module. This design strategically aims to discern and capture nuanced patterns associated with peak fluctuations in seasonal time-series. This sophisticated architecture excels at capturing correlations between local and global features. To substantiate the efficacy of our proposed network structure, we conducted experiments by replacing SDNet with a self-attention module in the training of MSSD. The subsequent results, detailed in Table ~\ref{tab:ablation_mssd}, unequivocally confirm the superior effectiveness of our novel SDNet framework in enhancing the modeling and prediction capabilities compared to the alternative self-attention module.

In addition, we replaced the original convolutional module in MICN with the convolutional module proposed in this paper, and the result is shown in Table ~\ref{tab:ablation_mssd}. The effectiveness of the module proposed in this paper can be seen in Table ~\ref{tab:ablation_micn}.

\begin{table}[htbp]
    \centering
    \resizebox{\linewidth}{!}{
    \begin{tabular}{lcccccccccccccc}
    \toprule
    Models & Metrics & \multicolumn{4}{c}{CAISO} & \multicolumn{4}{c}{Electricity} & \multicolumn{4}{c}{Traffic} \\
    \cmidrule(lr){3-6} \cmidrule(lr){7-10} \cmidrule(lr){11-14}
    & & 48 & 96 & 336 & 720 & 48 & 96 & 336 & 720 & 48 & 96 & 336 & 720 \\
    \midrule
    MSSD-SDNet & MSE & \textbf{0.091} & \textbf{0.101} & \textbf{0.097} & \textbf{0.093} & 0.076 & 0.061 & \textbf{0.068} & \textbf{0.064} & \textbf{0.146} & \textbf{0.108} & \textbf{0.146} & \textbf{0.17} \\
               & MAE & \textbf{0.25}  & \textbf{0.262} & \textbf{0.258} & \textbf{0.255} & 0.224 & 0.203 & \textbf{0.214} & \textbf{0.207} & \textbf{0.325} & \textbf{0.263} & \textbf{0.327} & \textbf{0.346} \\
    \midrule
    MSSD-Auto correlation & MSE & 0.108 & 0.132 & 0.121 & 0.104 & \textbf{0.072} & \textbf{0.058} & 0.079 & 0.081 & 0.157 & 0.128 & 0.165 & 0.178 \\
                          & MAE & 0.264 & 0.303 & 0.275 & 0.269 & \textbf{0.191} & \textbf{0.193} & 0.227 & 0.252 & 0.341 & 0.324 & 0.356 & 0.359 \\
   \midrule
    MSSD-self attention & MSE & 0.123 & 0.146 & 0.17  & 0.154 & 0.104 & 0.117 & 0.095 & 0.094 & 0.199 & 0.204 & 0.22  & 0.215 \\
                         & MAE & 0.346 & 0.391 & 0.384 & 0.372 & 0.297 & 0.306 & 0.299 & 0.276 & 0.453 & 0.472 & 0.487 & 0.468 \\
    \bottomrule
    \end{tabular}
    }
    \caption{Ablation of Local-global module in the MSSD.}
    \label{tab:ablation_mssd}
\end{table}

\begin{table}[htbp]
    \centering
    \resizebox{\linewidth}{!}{
    \begin{tabular}{lcccccccccccccc}
    \toprule
    Models & Metrics & \multicolumn{4}{c}{CAISO} & \multicolumn{4}{c}{Electricity} & \multicolumn{4}{c}{Traffic} \\
    \cmidrule(lr){3-6} \cmidrule(lr){7-10} \cmidrule(lr){11-14}
    & & 48 & 96 & 336 & 720 & 48 & 96 & 336 & 720 & 48 & 96 & 336 & 720 \\
    \midrule
    MICN-SDNet & MSE & \textbf{0.107} & 0.108 & \textbf{0.119} & 0.127 & \textbf{0.299} & \textbf{0.304} & \textbf{0.319} & \textbf{0.357} & 0.158 & 0.16 & \textbf{0.165} & 0.189 \\
               & MAE & \textbf{0.249} & 0.271 & \textbf{0.287} & \textbf{0.284} & 0.347 & 0.405 & \textbf{0.399} & \textbf{0.437} & \textbf{0.314} & 0.246 & \textbf{0.242} & 0.267 \\
   \midrule
    MICN & MSE & 0.112 & 0.106 & 0.121 & 0.124 & 0.304 & 0.31 & 0.323 & 0.364 & 0.167 & 0.158 & 0.165 & 0.182 \\
         & MAE & 0.257 & 0.269 & 0.296 & 0.285 & 0.354 & 0.398 & 0.413 & 0.449 & 0.329 & 0.241 & 0.243 & 0.264 \\
    \bottomrule
    \end{tabular}
    }
    \caption{Comparison of Models: MICN-SDNet vs. MICN}
    \label{tab:ablation_micn}
\end{table}

\textbf{Robustness analysis:} 
We used a direct noise injection approach to evaluate the robustness of our model. Using this method, we introduce varying degrees of perturbation $\zeta$ to the training data and subsequently record the Mean Absolute Error (MAE) index. The findings, detailed in Table 6, reveal a slight increase in the predicted MAE index with escalating perturbation $\zeta$ ratios. Significantly, these results highlight the robust nature of MSSD, demonstrating its resilience, especially up to a 20\% perturbation threshold. This resilience signifies MSSD's proficiency in effectively handling mildly noisy data and excelling in accurately forecasting anomalous patterns within seasonal time series.

\begin{table}[htbp]
    \centering
    \resizebox{\linewidth}{!}{
    \begin{tabular}{lccccc}
    \toprule
    Missing & Metrics & \multicolumn{4}{c}{Electricity} \\
    \cmidrule(lr){2-5}
    & & 48 & 96 & 336 & 720 \\
    \midrule
    0\% & MSE & 0.076 & 0.061 & 0.068 & 0.064 \\
    & MAE & 0.224 & 0.203 & 0.214 & 0.207 \\
    \midrule
    5\% & MSE & 0.092 & 0.104 & 0.079 & 0.081 \\
    & MAE & 0.276 & 0.225 & 0.227 & 0.203 \\
    \midrule
    10\% & MSE & 0.121 & 0.151 & 0.137 & 0.129 \\
    & MAE & 0.345 & 0.372 & 0.364 & 0.331 \\
    \midrule
    20\% & MSE & 0.174 & 0.182 & 0.185 & 0.174 \\
    & MAE & 0.409 & 0.421 & 0.42 & 0.404 \\
    \bottomrule
    \end{tabular}
    }
    \caption{Robustness analysis of forecasting results.}
    \label{tab:missing_value}
\end{table}

\begin{figure}[h] 
  \centering
  \includegraphics[width=0.5\textwidth]{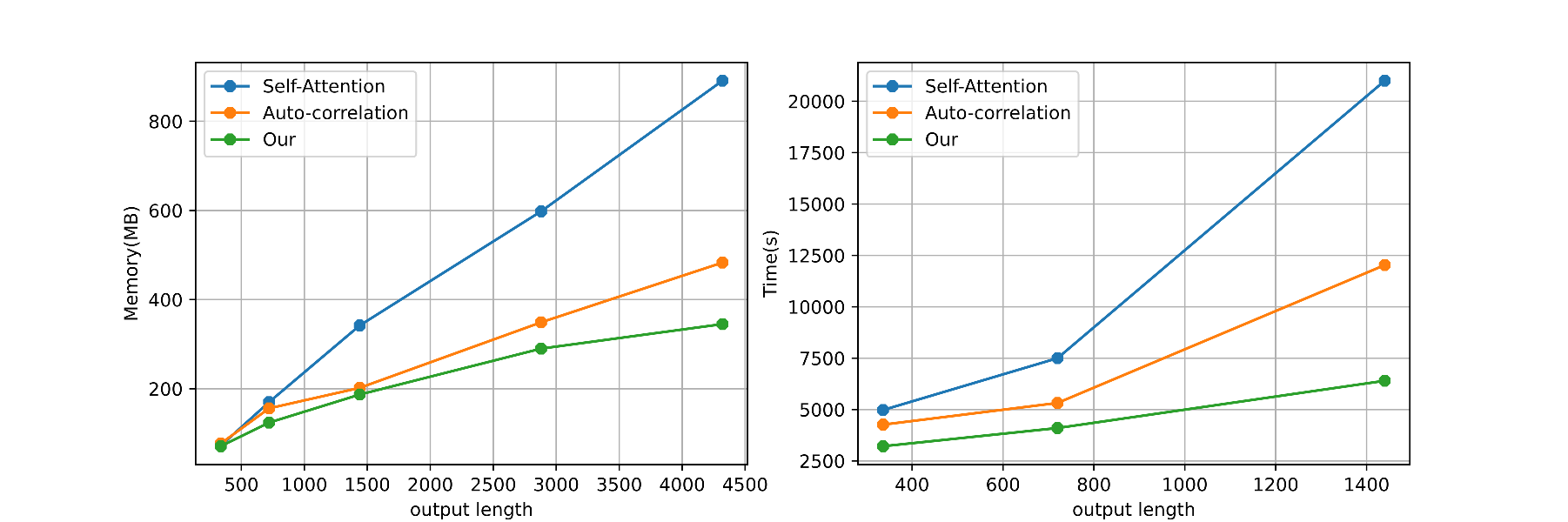} 
  \captionsetup{justification=centering}
  \caption{ Efficiency Analysis(local-global module vs self-attention, auto-correlation).}
  \label{fig:efficiency}
\end{figure}

\textbf{Efficiency analysis:} 
With increasing input length, our module exhibits a significant advantage, demanding less time and memory compared to both self-attention and Auto-correlation counterparts. This efficiency becomes especially pronounced with the increasing input length, confirming the scalability and resource efficiency of our module. The results emphasize its computational superiority, indicating that our module provides a more streamlined and resource-efficient solution, well-suited for managing extended input sequences with improved speed and reduced memory requirements compared to self-attention and Auto-correlation approaches.Fig ~\ref{fig:efficiency}

\textbf{Input length:} In the time-series prediction tasks, the input length significantly influences the algorithm's ability to leverage historical information for effective modeling. Models with a strong capacity to capture extended time dependencies are expected to perform well as the input length increases. To assess our model's performance, we conducted experiments with varying input lengths while keeping the prediction length consistent. Significantly, with the increasing input length, the performance of transformer-based models showed instability attributed to recurrent short-term patterns. In contrast, MSSD demonstrated a gradual and consistent improvement in predictive performance with increasing input length, highlighting its proficiency in capturing and utilizing long-term time dependencies for enhanced information extraction.
\begin{figure}[h] 
  \begin{minipage}[t]{\textwidth}
    \includegraphics[width=0.5\textwidth]{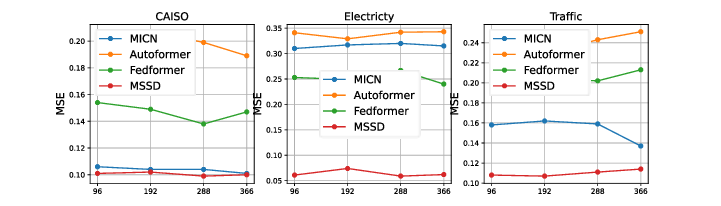}\\
  \end{minipage}
  \hfill 
  \begin{minipage}[t]{\textwidth}
    \includegraphics[width=0.5\textwidth]{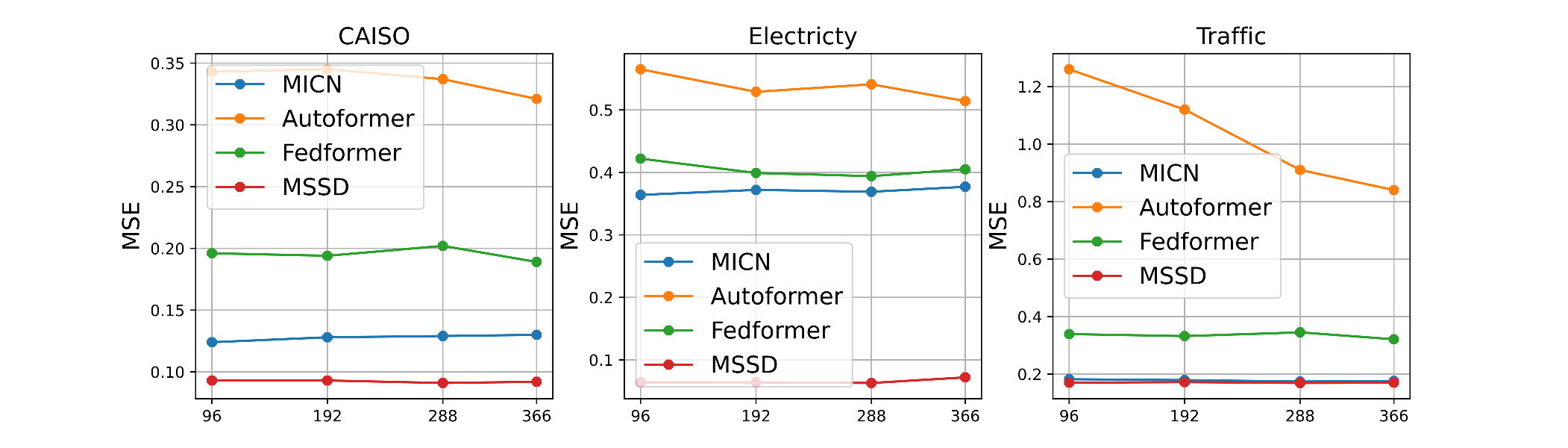}
  \end{minipage}
  \caption{The MSE results with different input lengths $O \in \{24, 720\}$.}
  \label{fig:input}
\end{figure}

\begin{figure}[h] 
  \centering
  \includegraphics[width=0.5\textwidth]{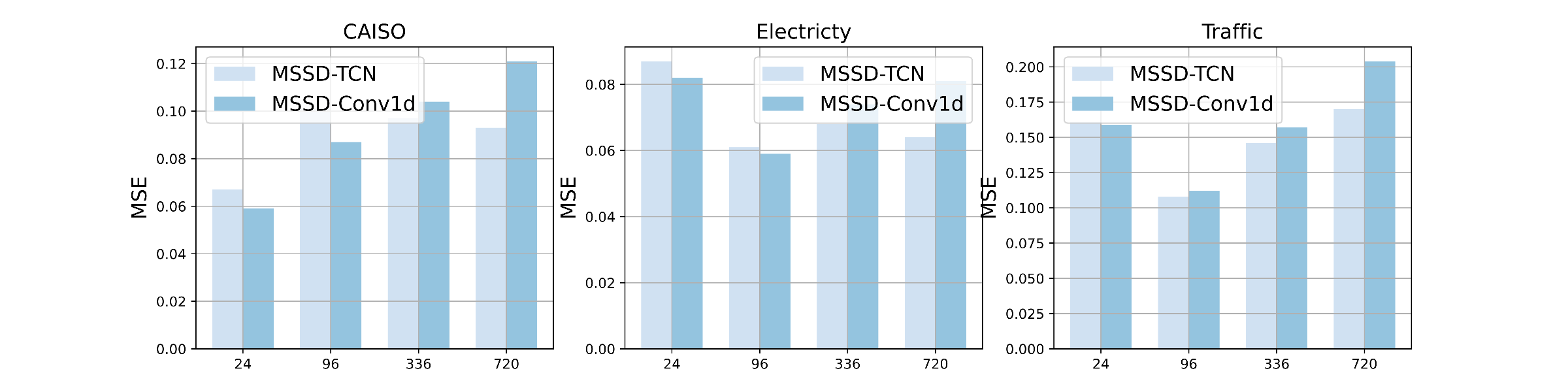} 
  \caption{ The MSE results with TCN vs Conv1d  $O \in \{24,96,336,720\}$.}
  \label{fig:tcn}
\end{figure}

\textbf{Casual Conv Vs. Conv1d:}
In this section, we conduct a comparative analysis involving TCN (Temporal Convolutional Network), known for its representation of causal convolution with an inflationary effect. For this comparison, we replace components of TCN with an equivalent 1-dimensional convolution and evaluate the outcomes for various output lengths: 24, 96, 336, and 720. As depicted in Fig.~\ref{fig:tcn}, upon close examination of the results, a clear pattern emerges—Conv1d consistently outperforms TCN in the context of short-term prediction tasks. In contrast, when faced with the challenges of long-term forecasting, TCN demonstrates a more robust and superior modeling capability.

\begin{figure}[h] 
  \centering
  \includegraphics[width=0.5\textwidth]{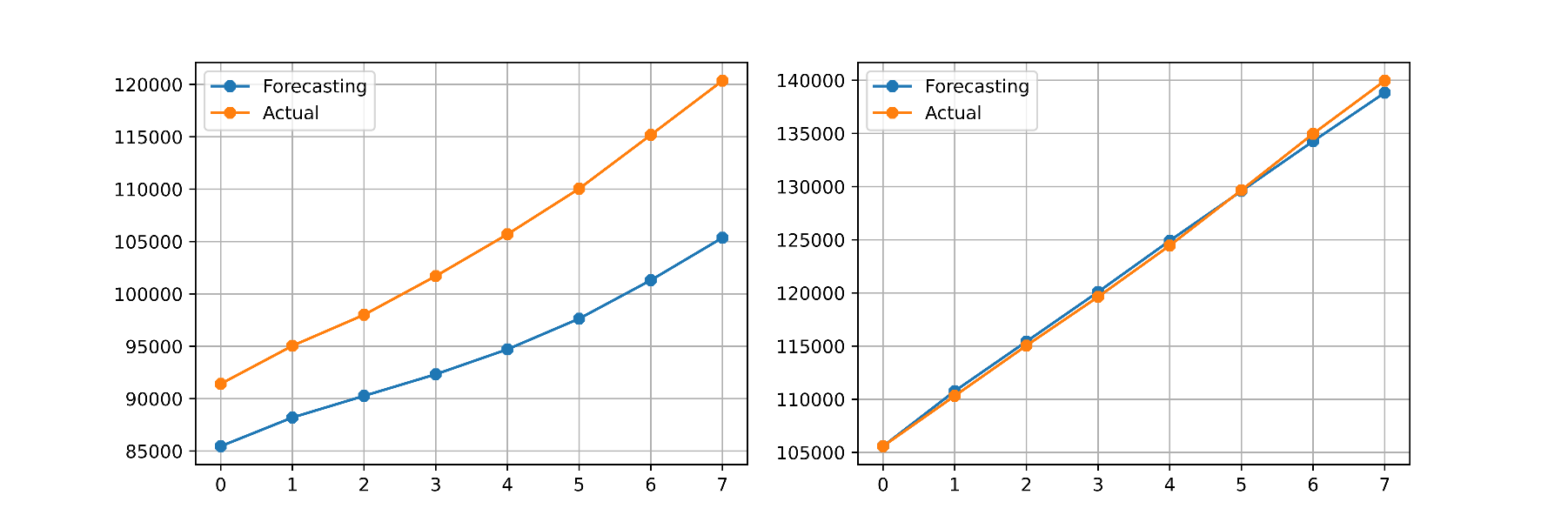} 
  \caption{Ascending part forecasting for MSSD.}
  \label{fig:linear}
\end{figure}

\textbf{Input visualization:}
One-dimensional convolution effectively captures local features of the peak component, while causal convolution adeptly captures global features. Fig \ref{fig:linear} illustrates linear regression for predicting the ascending component. It is evident that linear regression effectively models both the ascending and descending components.

\begin{figure}[h] 
  \centering
  \includegraphics[width=0.5\textwidth]{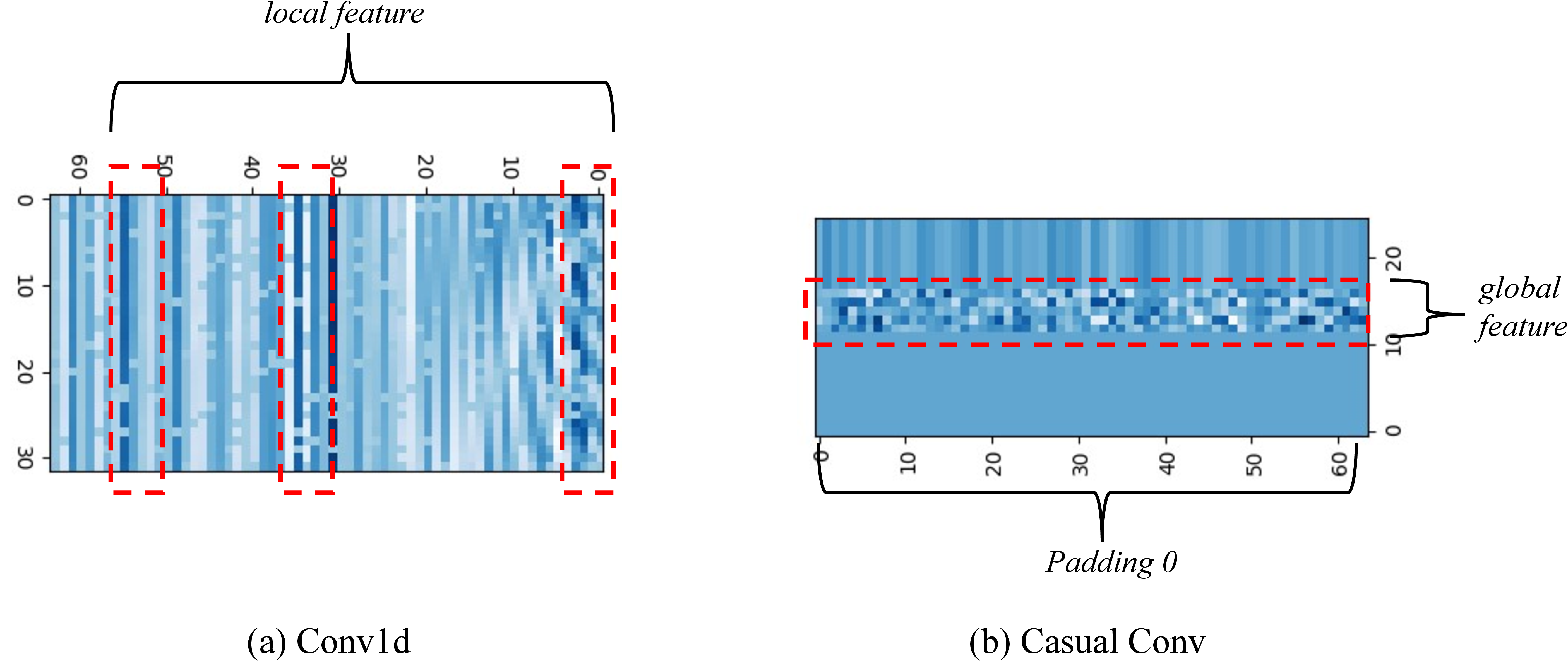} 
  \caption{Feature of Conv1d and Casual Conv}
  \label{fig:conv_weight}
\end{figure}

\begin{figure}[h] 
  \centering
  \includegraphics[width=0.5\textwidth]{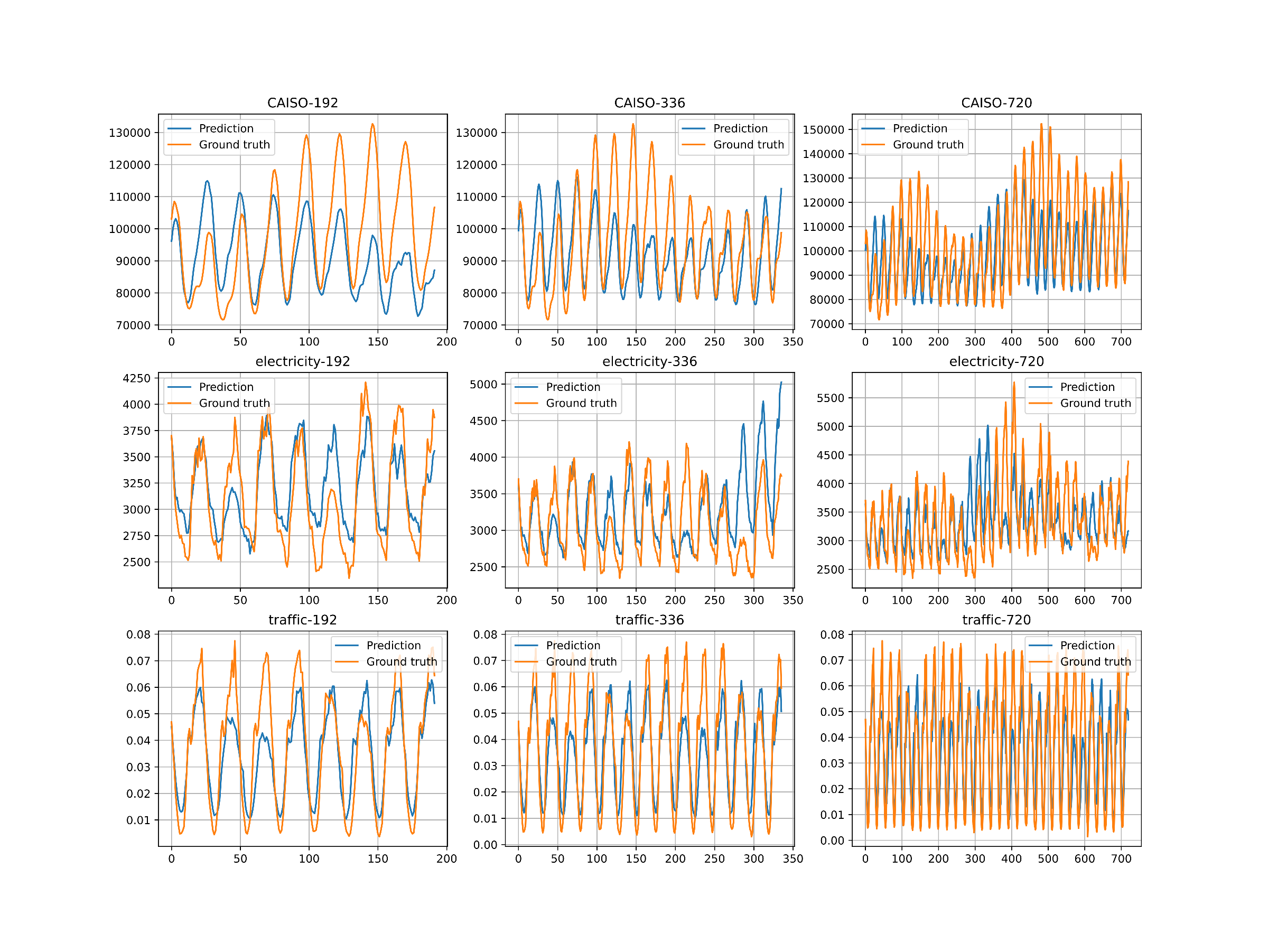} 
  \caption{Visualization of the Electricity, Traffic, and CAISO datasets (prediction length = 192, 336, 720).}
  \label{fig:long}
\end{figure}

\section{Conclusion}
In this paper, we present MSSD, a highly specialized convolution-based prediction framework meticulously crafted to address the nuances of seasonal time-series data. MSSD stands out by predicting the rising, peak, and falling segments of seasonal time series individually, offering a comprehensive understanding of the underlying patterns. Across three diverse real-world datasets, MSSD consistently demonstrates state-of-the-art performance, showcasing its efficacy and reliability in forecasting seasonal trends. Notably, our approach to peak prediction involves the strategic use of diverse convolution operations, marking a departure from conventional methods. Through rigorous comparative experiments, we establish that our convolution-based prediction module surpasses the performance of the attention mechanism, the most popular technique in time-series analysis.

A key feature of MSSD is its simultaneous partitioning of seasonal time series into rising, peak, and falling segments. This segmentation not only contributes to the model's accuracy but also significantly enhances interpretability. By addressing each phase of the seasonal pattern independently, MSSD provides valuable insights into the distinct characteristics of the data.

At present, MSSD is mainly aimed at periodic time-series, and there is no obvious periodic time-series data such as Exchange and Weather, so it cannot achieve a more accurate prediction. In future work, we plan to make targeted improvements to MSSD to accommodate a wider range of time-series data characteristics. Especially for volatile data such as trading and weather, we will explore new model structures and parameter adjustments to improve forecast accuracy and robustness. This will involve a deeper level of model tuning to ensure that the MSSD exhibits superior performance across different types of time-series data.

\bibliographystyle{IEEEtran}
\bibliography{reference.bib}
\end{document}